\pdfoutput=1
\documentclass[11pt]{article}

\usepackage[dvipsnames, table]{xcolor}

\usepackage[]{naacl2021}
\usepackage{amsfonts}
\usepackage{bbm}
\usepackage{graphicx}
\usepackage{amsmath}
\usepackage{times}
\usepackage{tcolorbox}
\usepackage{latexsym}
\usepackage{booktabs}
\usepackage{multirow}
\usepackage[T1]{fontenc}
\usepackage[utf8]{inputenc}
\usepackage{microtype}

\definecolor{lightred}{RGB}{240, 180, 180}
\definecolor{lightgreen}{RGB}{150, 243, 150}

\newcommand{\colorcellgreen}[3]{%
  \pgfmathsetmacro{\perc}{(1-(#1-#2)/(#3-#2))*100}%
  \edef\cellcolorcmd{\noexpand\cellcolor{white!\perc!lightgreen}}%
  \cellcolorcmd
}

\newcommand{\colorcellred}[3]{%
  \pgfmathsetmacro{\perc}{((#1-#2)/(#3-#2))*100}%
  \edef\cellcolorcmd{\noexpand\cellcolor{white!\perc!lightred}}%
  \cellcolorcmd
}

\title{Counting Ability of Large Language Models and Impact of Tokenization}

\author{
Xiang Zhang$^{1}$\thanks{$^{*}$Equal contribution.} \qquad Juntai Cao$^{1*}$ \qquad Chenyu You$^{2}$ \\ 
$^{1}$ University of British Columbia \\
$^{2}$ Stony Brook University \\ 
\texttt{xzhang23@ualberta.ca, jtcao7@cs.ubc.ca, chenyu.you@stonybrook.edu}}

\begin{document}
\maketitle
\begin{abstract}
Transformers, the backbone of modern large language models (LLMs), face inherent architectural limitations that impede their reasoning capabilities. Unlike recurrent networks, Transformers lack recurrent connections, confining them to constant-depth computation. This restriction places them in the complexity class TC$^0$, making them theoretically incapable of solving tasks that demand increasingly deep reasoning as input length grows. Counting, a fundamental component of many reasoning tasks, also requires reasoning depth to grow linearly to be performed inductively. While previous studies have established the upper limits of counting ability in Transformer-based expert models (i.e., models specifically trained for counting tasks), these findings do not directly extend to general-purpose LLMs due to differences in reasoning mechanisms. Recent work has highlighted how Chain of Thought (CoT) reasoning can help alleviate some of the architectural limitations of Transformers in counting tasks. However, little attention has been paid to the role of tokenization in these models. Unlike expert models that often use character-level tokenization, LLMs typically rely on byte-level (BPE) tokenizers, which fundamentally alters the way reasoning is processed. Our work investigates the impact of tokenization on the counting abilities of LLMs, uncovering substantial performance variations based on input tokenization differences. We provide both theoretical and experimental analyses, offering insights into how tokenization choices can undermine models' theoretical computability, thereby inspiring the design of new tokenization methods to enhance reasoning in LLMs. All code, prompts, and experiment logs, API returns are  released on \href{https://github.com/juntaic7/Impact-of-Tokenization-in-the-Counting-Ability-of-Language-Models}{ GitHub}.
\end{abstract}

\section{Introduction}
\textit{Counting}, a fundamental component of most complex reasoning tasks, has been extensively studied across various disciplines for decades~\cite{boolos2002computability,wynn1990children,de1964polya}. In particular, computer scientists have explored the circuit complexity of counting~\cite{jerrum1995computational}, the needed capabilities of computing machines~\cite{fischer1968counter} to perform counting, and how counting relates to more complex tasks within the framework of computability theory~\cite{ibarra2002counter}.
Basic counting (counting from \texttt{1} to $\texttt{n}$) requires a depth complexity, the number of sequential computation steps, that grows with input length (non-constant). This requirement, grounded in computability theory~\cite{fischer1968counter}, serves as a theoretical constraint for any computational machine to solve this task, from simple state machines~\cite{cooper2017computability} to neural networks~\cite{lecun2015deep,jin2024contranovo}. Previous research has thoroughly examined the computability, the ability to solve tasks of a given complexity, of different types of neural networks~\cite{deletang2022neural,zhang2024autoregressivechainthought}. Both theoretical~\cite{raghu2017expressive} and experimental~\cite{deletang2022neural} findings have shown that \textit{counting} requires an equivalent or higher level of computability than what LSTM and RNN provide~\cite{deletang2022neural}, with the former empirically aligning with the capabilities of $\texttt{k}$-counter machines~\cite{deletang2022neural}.

\begin{figure}[t!]
    \centering
    \includegraphics[width=\linewidth]{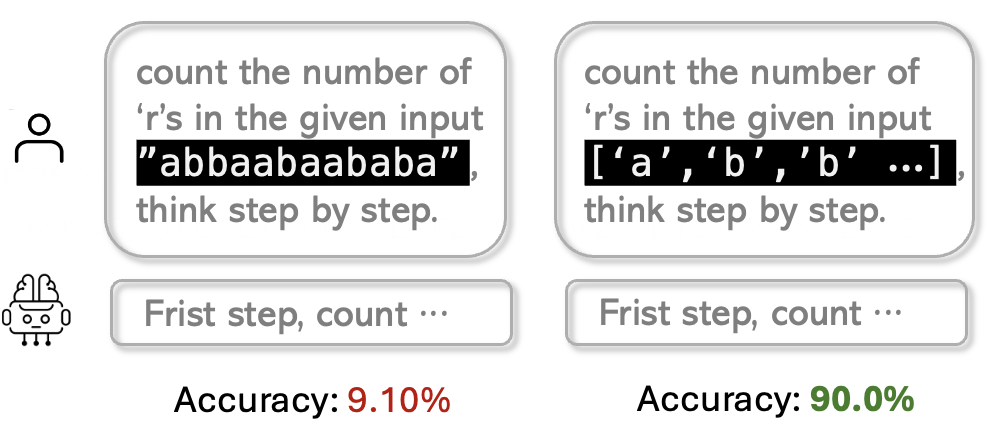}
    \vspace{-0.5em}
    \caption{Experimental results on average counting accuracy based on different tokenization choices, using GPT-4o mini. Our approach treats the model as a \textit{black-box}, manipulating BPE tokenizers to function differently through carefully engineered string formats.}
    \label{fig:1}
    \vspace{-1.5em}
\end{figure}
Transformers~\cite{vaswani2017attention}, both autoregressive~\cite{gregor2014deep,achiam2023gpt,zhang-etal-2023-dont} (GPT-like) and non-autoregressive~\cite{devlin2018bert,liu2022character,zhang2024pi} (BERT-like), are constrained to the \textit{constant-depth} computation~\cite{zhang2024autoregressivechainthought,deletang2022neural,li2024chain}. Neural networks perform all computations and reasoning in a latent space, $\mathcal{H}$, where the hidden state $\mathbf{h}$ serves as the medium for computational state storage and information processing. Unlike recurrent architectures, where $\mathbf{h}_\texttt{t} = g_\theta (\mathbf{h}_\texttt{t-1})$ allows hidden states to evolve over time, Transformers update $\mathbf{h}$ sequentially only through layers rather than time steps. Consequently, Transformer with a fixed number of layers can only process (or reason with) $\mathbf{h}$ a constant number of times, placing them at the lower end of the Chomsky computational hierarchy~\cite{deletang2022neural}. This architectural limitation prevents Transformer-based models—from small expert model to LLMs—from performing \textit{counting} tasks relying solely on their internal hidden states computation.

Chain of Thought~\cite{wei2022chain} reasoning has revolutionized the mechanism by which reasoning can be performed. Specifically, it \textit{extends} reasoning from the latent space $\mathcal{H}$ to the text space $\mathcal{O}$~\cite{zhang2024autoregressivechainthought}, using sequences of natural language (referred to as ``\textit{Thought}'') to relay the computations of $\mathbf{h}$ in the absence of recurrence. As a result, higher-complexity tasks, such as counting, become feasible. Several theoretical works~\cite{li2024chain,zhang2024autoregressivechainthought,feng2024towards} have proven that CoT-augmented LLMs have an \textit{upper-bound} (under ideal assumptions) for performing computational tasks of arbitrary complexity, including \textit{counting}.

Despite extensive analysis and theoretical guarantees on the upper bound for counting ability, a significant gap remains in practical performance~\cite{zhang2024autoregressivechainthought,chang2024language}. As modern LLMs scale from millions to billions of parameters~\cite{achiam2023gpt}, improvements in counting ability have been minimal—GPT-4, for example, struggles with simple tasks like counting the number of "r"s in a word. While recent research has examined factors such as training data selection~\cite{allen2023physics,yin2023ttida} and positional encoding~\cite{chang2024language}, one of the most fundamental aspects, tokenization, has received relatively little attention. Specifically, modern byte pair encoding (BPE)~\cite{sennrich2015neural} tokenization for LLMs groups multiple characters into a single token for efficiency, which will inadvertently degrade performance in arithmetic tasks due to information loss during the tokenization process.

In this work, we thoroughly investigate how the choice of tokenization can significantly undermine the theoretical counting ability of neural models. We adopt a model-agnostic approach, allowing for the analysis of closed-source LLMs with unknown tokenizers. Through extensive counting experiments using CoT (theoretically proven to be Turing complete under ideal conditions), we demonstrate that the correct tokenization choice is crucial for fully out-bringing a model's theoretical counting ability and bridging the gap between theory and practice. Otherwise, accuracy drops of up to $80\%$ can be observed. Additionally, our experiments reveal that the impact of tokenization varies across models, with certain tokens proving more sensitive in counting tasks, even when the nature of the task remains unchanged.


\section{Neural Networks and Counting: A Revisit}
\textbf{Training neural networks for counting.} Counting has been extensively studied in neural networks (NNs) as it is a fundamental skill required for more advanced tasks~\cite{chang2024language}. Since multi-layer perceptrons (MLPs)~\cite{rosenblatt1958perceptron} can only handle fixed-length inputs, which contradicts the nature of counting, early NN training for counting tasks began with recurrent neural networks (RNNs). \citet{rodriguez1999recurrent} trained an early RNN to recognize the \textit{regular} language $\texttt{a}^\texttt{n}\texttt{b}^\texttt{n}$, which required the network to count the number of \texttt{a}s and \texttt{b}s in the input string. Of the 50 networks trained, 8 successfully learned the task and generalized to longer strings, demonstrating the counting ability of RNNs.
\citet{suzgun2019lstmnetworksperformdynamic} trained LSTM for counting in the context of \textit{bracket pairing}, showing that LSTMs could perform \textit{dynamic counting} by maintaining separate counters (represented by different types of brackets) using their memory (cell state) and gating mechanisms—something RNNs were unable to achieve. More recently, \citet{deletang2022neural} systematically explored counting in mainstream NNs, including RNNs, LSTMs, and Transformers. Their experiments showed that Transformers could not perform counting, while LSTMs exhibited computability aligned with \textit{counter machines}.
\citet{chang2024language} extended the analysis of counting performance in Transformers to other modern architectures such as Mamba~\cite{gu2023mamba} and RWKV~\cite{peng2023rwkv}, demonstrating that these models also struggle with counting tasks outside their training distribution, performing worse than RNNs empirically.

\noindent\textbf{Analyzing neural networks' ability to count.} Counting falls within the domain of computability, which studies a machine's ability to perform tasks of varying complexity. Researchers have made efforts to theoretically understand the upper bounds of each type of neural network from a computability perspective. \citet{sperduti1997computational} and \citet{tivno1998finite} were among the first to theoretically align RNNs with deterministic finite automata (DFA), showing that, under proper assumptions, they are equivalent. In this context, RNNs act as one-finite-counter machines, capable of performing basic counting tasks (e.g., parity checks).
Recent studies~\cite{weiss2018practicalcomputationalpowerfinite,ackerman2020survey} offer further insights and more rigorous proofs on the computability of RNNs and LSTMs, theoretically confirming earlier experimental results on their counting abilities. However, Transformers, lacking inherent recurrence, are limited to TC$^{0}$~\cite{li2024chain} in terms of inductive reasoning. As demonstrated in recent theoretical work~\cite{sanford2024transformers,li2024chain,deletang2022neural}, this places Transformers at the lowest level of the Chomsky hierarchy, making it impossible for them to learn counting tasks unless certain inductive bias is introduced\cite{chang2024language}.
\begin{figure*}[t!]
    \centering
    \includegraphics[width=\linewidth]{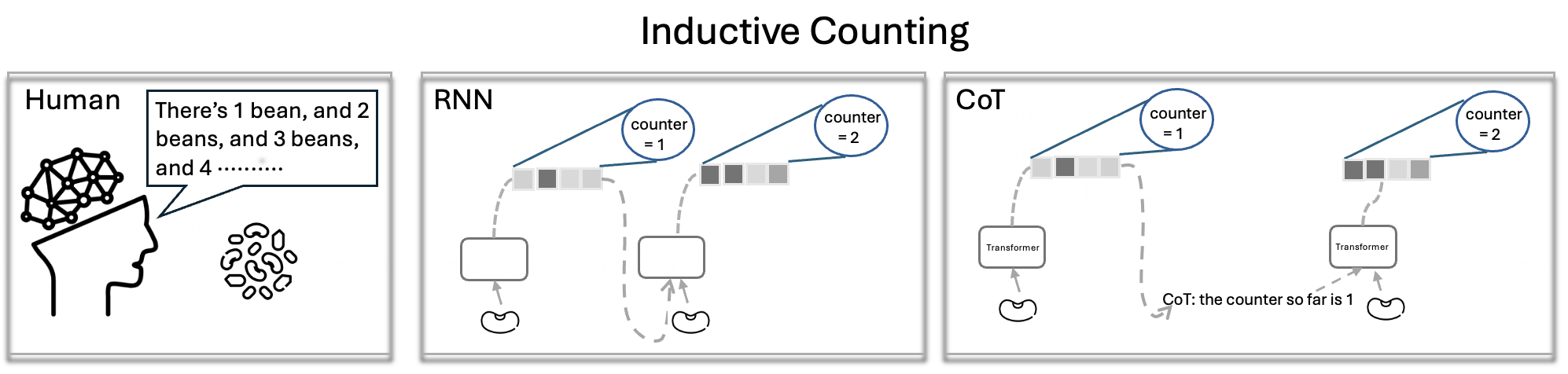}
    \vspace{-1.5em}
    \caption{Illustration of inductive counting as performed by humans, RNNs, and LLMs with CoT, respectively. }
    \label{fig:inductive}
    \vspace{-1.em}
\end{figure*}
\section{CoT + Ideal Assumption = Complete Counting Ability}
Mainstream LLMs use the Transformer architecture as their backbone~\cite{bai2023qwen,touvron2023llama,achiam2023gpt}, and, as a result, inherit the computability limitations of Transformers, particularly in tasks like counting. Given these constraints, it's unsurprising that Transformers struggle with counting tasks. However, Chain of Thought (CoT) reasoning has revolutionized how Transformers reason, offering new possibilities counting task. In this section, we demonstrate \textit{what} counting entails and \textit{why} Transformers alone cannot effectively solve counting tasks. We then show \textit{how} CoT can achieve perfect counting accuracy under ideal assumptions.

\subsection{Inductive Counting and Computation}
Counting is inherently inductive in both human cognition and computability theory~\cite{borodin1989two}. To count from $\texttt{1}$ to $\texttt{n}$, one must start at $\texttt{1}$, then proceed step by step—counting to $\texttt{2}$ before reaching $\texttt{3}$, and so on. Humans typically count inductively, except for very small numbers (1-3), where we rely on memorization and can immediately recognize the total at a glance. For larger quantities, like a pile of apples, we engage in iterative inductive counting from 1 to $\texttt{n}$ (Figure~\ref{fig:inductive}).
In computability, a state machine processes an input string token by token, transitioning its internal state after each token to keep track of the count. This step-by-step state transition aligns with the inductive nature of counting. Similarly, neural networks perform counting internally through their hidden state, $\mathbf{h}$, which serves as the location for reasoning and intermediate information storage (e.g., counter storage) (Figure~\ref{fig:inductive}. In RNNs, as $\mathbf{h}$ is updated with each new input token $\text{x}_\texttt{t}$ via recurrent connections, the counter can be inductively updated at every step (Figure~\ref{fig:inductive}). The inductive bias of RNNs allows $\mathbf{h}$ to act as the smallest \textit{reasoning unit}, incrementing the stored counter with each update.

However, Transformers can only sequentially process $\mathbf{h}$ a fixed number of times~\cite{li2024chain,sanford2024transformers,zhang2024autoregressivechainthought}, limited by their number of layers. For instance, when counting the number of $\texttt{a}$s in the input string $\texttt{aababaa}$, the Transformer initializes its counter in the latent representation $\mathbf{h}$. While each layer captures substantial computation through matrix operations ($\mathbf{W}\mathbf{X}$), because $\mathbf{h}$ is computed in parallel across positions $\mathrm{t}$, it lacks the depth (sequential counting) needed for inductive counting. The counter is only updated when the hidden state $\mathbf{h}$ is sequentially passed from one layer to the next, limiting the model to a fixed number of counting steps.
Some theoretical work~\cite{chiang2022overcoming} suggests that Transformers can perform counting with hard-coded weights, but this approach is not \textit{inductive}. Instead, it treats each bit in $\mathbf{h}$ and $\mathbf{W}$ as the smallest unit of reasoning (in contrast to treating the entire $\mathbf{h}$ as a unit in RNNs) and performs \textit{circuit-level} computations. This relies on the bit-level dependencies within matrix multiplication (e.g., $\mathrm{w}_\texttt{1}\mathrm{x}_\texttt{1} + \mathrm{w}_\texttt{2}\mathrm{x}_\texttt{2} + \cdots + \mathrm{w}_\texttt{n}\mathrm{x}_\texttt{n}$), breaking the abstraction of neural computation and delving into circuit-level operations.
As a result, training a standard Transformer, as demonstrated by various experimental studies~\cite{chang2024language,deletang2022neural}, consistently fails to learn counting in a generalizable manner, unlike RNNs. Transformers do not converge on these "theoretically possible" counting solutions unless additional inductive biases are introduced, such as specific positional encodings or architectural modifications~\cite{chang2024language,fan2020addressing}.

\subsection{CoT: Sky is the Limit}
As Transformer-based models, both LLMs and expert models, struggle to count effectively using only their internal reasoning state~\cite{chang2024language,deletang2022neural} (also demonstrated in experiments below), Chain of Thought (CoT)~\cite{wei2022chain} shifts the reasoning required for inductive counting into the text space~\cite{li2024chain,zhang2024autoregressivechainthought}. Instead of simply outputting a final counting result $\mathrm{y}$ after processing an input sequence $\mathbf{x} = (\mathrm{x}_\texttt{1}, \mathrm{x}_\texttt{2}, \mathrm{x}_\texttt{3}, \cdots )$, LLMs are guided to output intermediate reasoning steps, in this case counter value, after processing each input unit $\mathrm{x}_\texttt{i}$.
Since internal reasoning via $\mathbf{h}$ can only handle a limited number of sequential counting steps, CoT allows LLMs to convert the latent counter information from $\mathbf{h}$ into a sequence of tokens $(\mathrm{o}_\texttt{1}, \mathrm{o}_\texttt{2}, \cdots, \mathrm{o}_\texttt{k})$, referred to as \textit{Thought}, which represents the counter value in text (Figure~\ref{fig:inductive}). During subsequent computations, these \textit{thought} tokens are encoded back into the latent space through the embedding layer, whose information forming the new $\mathbf{h}$ for the next step of reasoning (Figure~\ref{fig:inductive}).
In essence, CoT approximates the recurrent computation in RNNs—$\mathbf{h}_\texttt{t-1} \Rightarrow \mathbf{h}_\texttt{t}$—by using $\mathbf{h}_\texttt{t-1} \Rightarrow (\mathrm{o}_\texttt{1}, \mathrm{o}_\texttt{2}, \cdots, \mathrm{o}_\texttt{k}) \Rightarrow \mathbf{h}_\texttt{t}$, where $(\mathrm{o}_\texttt{1}, \mathrm{o}_\texttt{2}, \cdots, \mathrm{o}_\texttt{k})$ encodes information from $\mathbf{h}_\texttt{t-1}$. This enables infinite reasoning depth when the CoT can be extended indefinitely (ideal assumption). Consequently, CoT allows counting to be performed iteratively and inductively, using  constant cycle of text-vector conversion to continuously update the counter~\cite{zhang2024autoregressivechainthought}.

\section{Tokenization as a Black Box Model}
Even with CoT enabled, significant failures in counting are still observed in modern LLMs like GPT-4, which consistently make errors when counting letters in words as short as 3 to 10 characters such as \texttt{Strawberry}~(Table \ref{tab:main_result_gpt}). Given the number of layers in models of this size~\cite{touvron2023llama}, counting within 10 digits internally should be feasible. A key factor contributing to these errors is the tokenizer used, specifically byte-level Byte Pair Encoding (BPE)~\cite{sennrich2015neural}. BPE groups a certain number of characters into tokens, both within and between words, leading to a mismatch between the unit to be counted (e.g., letters) and the unit actually being processed (BPE tokens).
In this section, we explore the potential impact of tokenization on the counting abilities of LLMs and introduce our novel method for analyzing this effect, treating the model as a \textit{black box}.

\begin{figure*}[t!]
    \centering
    \includegraphics[width=\linewidth]{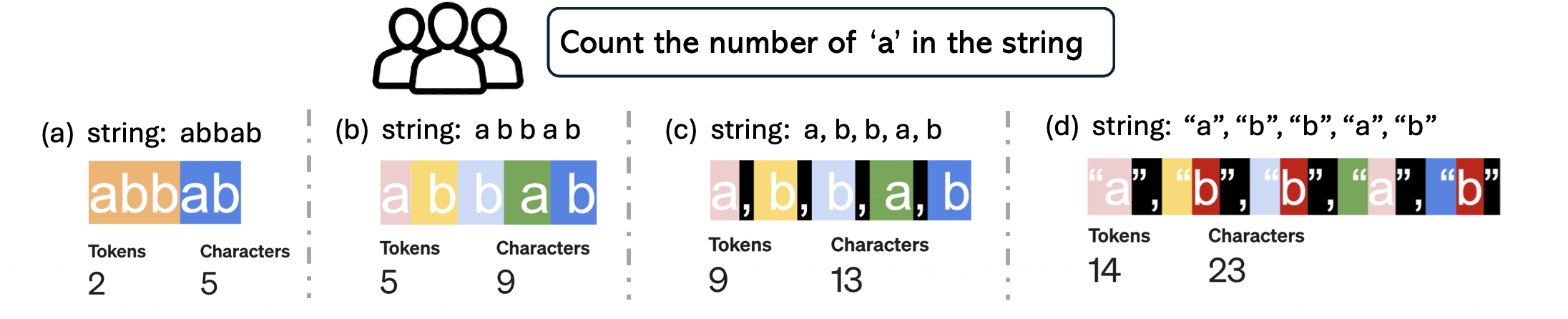}
    \vspace{-0.5em}
    \caption{Four types of string formatting used for counting tasks to manipulate tokenization in LLMs. Examples in the figure are tokenized using the GPT-4o tokenizer. Each string-token type is labeled as (a), (b), (c), and (d) in the diagram. Note that changing the format does not alter the fundamental nature or difficulty of the counting task. }
    \label{fig:3}
    \vspace{-1.5em}
\end{figure*}
\subsection{Imperfect Tokenizer + CoT < CoT Limit}
We introduce the concept of \textit{Token Awareness} in the context of LLMs. For any given token $\mathbf{t}$ (e.g., the token \texttt{pre}) produced after tokenization, each token has properties such as \texttt{how many "r"s are in this token}. Since LLMs are trained to predict entire \textit{tokens} rather than individual \textit{letters}, specific token properties may not be fully \textit{aware} by the model unless they can be inferred from the training corpus. During training, a token is mapped directly into its token embedding without additional information, meaning the model may not even know if this token contain any letters.

Now, consider a CoT-augmented LLM under ideal conditions, where it has perfect counting ability. If the counting is performed at a level that does not align with the tokenization—such as counting letters instead of BPE tokens—missing token awareness for the required granularity will lead to counting errors. For instance, if the $\texttt{i}$th token in the input string has token properties that the model is unaware of, counting at token $\texttt{i}$ will cause a shift in the total count. As an example, if the model is unaware of the number of $\texttt{r}$s in the first token $\texttt{str}$ when processing the word \texttt{strawberry}, the counting will fail right from the start. Thus, even with perfect \textit{Counting Ability}, tokenization becomes a barrier to successful task performance.

\subsection{Method}
Building on the theoretical analysis, our goal is to examine the practical counting abilities of LLMs, with a particular focus on the impact of tokenization—a crucial yet often under-explored factor in a model's counting potential. We propose a novel approach to studying the role of tokenization in LLMs' counting performance. Since many LLMs, along with their tokenization algorithms, are closed-source, we designed our approach to be model-agnostic, treating LLMs as black-box systems.

We begin by establishing universal assumptions based on typical LLM design in real-world models: (1) Consecutive letters of length 2-4 often form a single token; (2) Inputs beside delimiters like commas or spaces, are split into separate tokens. (3) After tokenization, delimiters (e.g., spaces, commas) are often merged with the preceding or following token. However, adding consecutive delimiters can prevent this merging and separate delimiter tokens from adjacent text tokens.

We validated our assumptions using multiple modern open-source LLMs or models with open tokenizers (Figure~\ref{fig:3} as well as Appendix Figure~\ref{fig:tokenall}). Based on these assumptions, we designed input instances for counting tasks to manipulate how tokenization is applied. To highlight the impact of tokenization on counting performance, we focused on letter-level counting, where the model needs token-level letter awareness to correctly update the counter in $\mathbf{h}$. It is important to note that the granularity of counting (letter vs. word level) does not change the fundamental nature of the counting task and should not affect the theoretical limits of counting ability.

To test how naive BPE tokenization affects counting, we designed counting instances as strings of consecutive letters, which are naturally tokenized by merging every 2-4 letters into a single token, based on assumption (1). Figure \ref{fig:3}.(a) illustrates the tokenization of such an example using the GPT-4o tokenizer. Formally, for a string $\mathbf{s}$ consisting of $\mathbf{n}$ letters, $(\mathrm{l}_\texttt{1}, \mathrm{l}_\texttt{2}, \cdots, \mathrm{l}_\texttt{n})$, the task is to count the total occurrences of a target letter $\mathrm{l}_\texttt{target}$:
\begin{equation}
    \mathbf{a} \in \mathbb{N}, \ \ \mathbf{a} = \sum_{i = \texttt{1}}^{\texttt{n}}\mathbbm{1}_{\{\mathrm{l}_\texttt{i} = \mathrm{l}_\texttt{target} \}} 
\end{equation}
where $\mathbf{a}$ is the answer (count), and $\mathbbm{1}$ is the indicator function, which returns \texttt{1} if the current letter matches the target letter.

\begin{table*}[t!]
\centering
\resizebox{\textwidth}{!}{
\begin{tabular}{rcccccc|cccccc}
\toprule
 & \multicolumn{6}{c|}{\textbf{Counting letter \texttt{a}}} & \multicolumn{6}{c}{\textbf{Counting letter \texttt{b}}} \\
\cmidrule(lr){2-7} \cmidrule(lr){8-13}
 \multirow{2}{*}{\textbf{String-Token Type \ \ \ }}& \multicolumn{2}{c}{\textbf{len $\in$ [10-20]}} & \multicolumn{2}{c}{\textbf{len $\in$ [20-30]}} & \multicolumn{2}{c|}{\textbf{len $\in$ [30-40]}} & \multicolumn{2}{c}{\textbf{len $\in$ [10-20]}} & \multicolumn{2}{c}{\textbf{len $\in$ [20-30]}} & \multicolumn{2}{c}{\textbf{len $\in$ [30-40]}} \\
\cmidrule(lr){2-3} \cmidrule(lr){4-5} \cmidrule(lr){6-7} \cmidrule(lr){8-9} \cmidrule(lr){10-11} \cmidrule(lr){12-13}
 & \textbf{no-CoT} & \textbf{CoT} & \textbf{no-CoT} & \textbf{CoT} & \textbf{no-CoT} & \textbf{CoT} & \textbf{no-CoT} & \textbf{CoT} & \textbf{no-CoT} & \textbf{CoT} & \textbf{no-CoT} & \textbf{CoT} \\
\midrule
\textbf{pure string BPE tokens (a)} &\colorcellgreen{30.10}{20}{99}30.10& \colorcellgreen{45.70}{20}{99}45.70& \colorcellred{15.10}{0}{20}15.10 & \colorcellred{9.10}{0}{20}9.10 & \colorcellred{6.40}{0}{20}6.40 & \colorcellred{2.00}{0}{20}2.00 & \colorcellgreen{33.20}{20}{99}33.20 & \colorcellgreen{47.70}{20}{99}47.70 & \colorcellred{14.00}{0}{20}14.00 & \colorcellred{9.40}{0}{20}9.40 & \colorcellred{3.80}{0}{20}3.80 & \colorcellred{2.70}{0}{20}2.70 \\
\textbf{\texttt{" "}-deliminated token (b)} & \colorcellgreen{46.20}{20}{99}46.20 &\colorcellgreen{58.40}{20}{99}58.40 & \colorcellred{16.10}{0}{20}16.10 & \colorcellgreen{24.90}{20}{99}24.90 & \colorcellred{7.50}{0}{20}7.50 & \colorcellred{10.90}{0}{20}10.90 & \colorcellgreen{45.90}{20}{99}45.90 & \colorcellgreen{63.70}{20}{99}63.70 & \colorcellred{17.60}{0}{20}17.60 & \colorcellgreen{34.00}{20}{99}34.00 & \colorcellred{5.60}{0}{20}5.60 & \colorcellred{18.60}{0}{20}18.60 \\
\textbf{\texttt{", "}-deliminated token (c)} & \colorcellgreen{56.00}{20}{99}56.00 & \colorcellgreen{55.40}{20}{99}55.40 & \colorcellred{19.40}{0}{20}19.40 & \colorcellgreen{38.60}{20}{99}38.60 & \colorcellred{10.20}{0}{20}10.20 & \colorcellgreen{28.10}{20}{99}28.10 & \colorcellgreen{63.60}{20}{99}63.60 & \colorcellgreen{69.30}{20}{99}69.30 & \colorcellgreen{32.80}{20}{99}32.80 & \colorcellgreen{56.10}{20}{99}56.10 & \colorcellred{13.90}{0}{20}13.90 & \colorcellgreen{42.30}{20}{99}42.30 \\
\textbf{precise-item token (d)} & \colorcellgreen{50.70}{20}{99}50.70 & \colorcellgreen{96.80}{20}{99}96.80& \colorcellred{15.80}{0}{20}15.80 & \colorcellgreen{81.60}{20}{99}81.60 & \colorcellred{7.90}{0}{20}7.90 & \colorcellgreen{56.10}{20}{99}56.10 & \colorcellgreen{58.30}{20}{99}58.30 & \colorcellgreen{96.50}{20}{99}96.50 & \colorcellgreen{30.20}{20}{99}30.20 & \colorcellgreen{90.00}{20}{99}90.00 & \colorcellred{12.60}{0}{20}12.60 & \colorcellgreen{70.80}{20}{99}70.80 \\
\bottomrule
\end{tabular}
}
\caption{Resulst of counting \texttt{a}s and \texttt{b}s in string consisting of letter \texttt{a} and \texttt{b}, using GPT-4o-mini API. Numbers indicate the average accuracy (\%) over 1000 random generated instances.  }
\label{tab:main_result_gpt}
\end{table*}

\begin{table*}[t]
\centering
\resizebox{\textwidth}{!}{
\begin{tabular}{rcccccc|cccccc}
\toprule
 & \multicolumn{6}{c|}{\textbf{Counting letter \texttt{e}}} & \multicolumn{6}{c}{\textbf{Counting letter \texttt{z}}} \\
\cmidrule(lr){2-7} \cmidrule(lr){8-13}
 \multirow{2}{*}{\textbf{String-Token Type \ \ \ }}& \multicolumn{2}{c}{\textbf{len $\in$ [10-20]}} & \multicolumn{2}{c}{\textbf{len $\in$ [20-30]}} & \multicolumn{2}{c|}{\textbf{len $\in$ [30-40]}} & \multicolumn{2}{c}{\textbf{len $\in$ [10-20]}} & \multicolumn{2}{c}{\textbf{len $\in$ [20-30]}} & \multicolumn{2}{c}{\textbf{len $\in$ [30-40]}} \\
\cmidrule(lr){2-3} \cmidrule(lr){4-5} \cmidrule(lr){6-7} \cmidrule(lr){8-9} \cmidrule(lr){10-11} \cmidrule(lr){12-13}
 & \textbf{no-CoT} & \textbf{CoT} & \textbf{no-CoT} & \textbf{CoT} & \textbf{no-CoT} & \textbf{CoT} & \textbf{no-CoT} & \textbf{CoT} & \textbf{no-CoT} & \textbf{CoT} & \textbf{no-CoT} & \textbf{CoT} \\
\midrule
\textbf{pure string BPE tokens (a)} & \colorcellgreen{26.60}{20}{99}26.60 & \colorcellgreen{55.20}{20}{99}55.20 & \colorcellred{19.80}{0}{20}19.80 & \colorcellred{12.20}{0}{20}12.20 & \colorcellred{11.40}{0}{20}11.40 & \colorcellred{2.10}{0}{20}2.10 & \colorcellgreen{31.10}{20}{99}31.10 & \colorcellgreen{59.10}{20}{99}59.10 & \colorcellred{11.70}{0}{20}11.70 & \colorcellgreen{22.10}{20}{99}22.10 & \colorcellred{4.60}{0}{20}4.60 & \colorcellred{7.30}{0}{20}7.30 \\
\textbf{\texttt{" "}-deliminated token (b)} & \colorcellgreen{41.00}{20}{99}41.00 & \colorcellgreen{52.90}{20}{99}52.90 & \colorcellgreen{23.90}{20}{99}23.90 & \colorcellgreen{28.20}{20}{99}28.20 & \colorcellred{13.00}{0}{20}13.00 & \colorcellred{16.00}{0}{20}16.00 & \colorcellgreen{45.30}{20}{99}45.30 & \colorcellgreen{63.99}{20}{99}63.90 & \colorcellred{16.60}{0}{20}16.60 & \colorcellgreen{46.20}{20}{99}46.20 & \colorcellred{6.80}{0}{20}6.80 & \colorcellgreen{29.50}{20}{99}29.50 \\
\textbf{\texttt{", "}-deliminated token (c)} & \colorcellgreen{45.50}{20}{99}45.50 & \colorcellgreen{64.20}{20}{99}64.20 & \colorcellgreen{27.40}{20}{99}27.40 & \colorcellgreen{44.20}{20}{99}44.20 & \colorcellred{18.00}{0}{20}18.00 & \colorcellgreen{27.60}{20}{99}27.60 & \colorcellgreen{56.20}{20}{99}56.20 & \colorcellgreen{73.60}{20}{99}73.60 & \colorcellgreen{28.20}{20}{99}28.20 & \colorcellgreen{55.60}{20}{99}55.60 & \colorcellred{13.90}{0}{20}13.90 & \colorcellgreen{41.90}{20}{99}41.90 \\
\textbf{precise-item token (d)} & \colorcellgreen{60.10}{20}{99}60.10 & \colorcellgreen{97.70}{20}{99}97.70 & \colorcellgreen{32.50}{20}{99}32.50 & \colorcellgreen{89.30}{0}{99}89.30 & \colorcellred{15.30}{0}{20}15.30 & \colorcellgreen{70.70}{0}{99}70.70 & \colorcellgreen{60.60}{20}{99}60.60 & \colorcellgreen{98.40}{20}{99}98.40 & \colorcellgreen{30.60}{20}{99}30.60 & \colorcellgreen{93.80}{20}{99}93.80 & \colorcellred{13.30}{0}{20}13.30 & \colorcellgreen{74.80}{20}{99}74.80 \\
\bottomrule
\end{tabular}
}
\caption{Resulst of counting \texttt{e}s and \texttt{z}s in string consisting of letter \texttt{e} and \texttt{z}, using GPT-4o-mini model.  }
\label{table:ez}
\end{table*}

As this approach inevitably merges consecutive letters, we introduce two alternative methods of inserting between-letter delimiters to simulate cases where \texttt{item-separated tokenization} is applied to the same counting instance. We use two types of delimiters, $\mathrm{d}_\texttt{1} = \texttt{" "}$ and $\mathrm{d}_\texttt{2} = \texttt{", "}$, as shown in Figure \ref{fig:3}.(b)-(c). According to assumptions (2) and (3), the resulting string $(\mathrm{l}_\texttt{1}, \mathrm{d}, \mathrm{l}_\texttt{2}, \mathrm{d}, \cdots, \mathrm{d}, \mathrm{l}_\texttt{n})$ manipulates the tokenizer to separate each item to be counted into its own token. However, the delimiter $\mathrm{d}$ and the letter $\mathrm{l}\texttt{i}$ will often be merged into a new token $\mathbf{t}$. Accurately counting the instance thus requires token awareness of the merged token with the delimiter.

\begin{table}[t]
\centering
\resizebox{0.47\textwidth}{!}{
\begin{tabular}{c|cccccc}
\toprule
\multirow{1}{*}{\textbf{string-token }} & \multicolumn{2}{c}{\textbf{len $\in$ [10-20]}} & \multicolumn{2}{c}{\textbf{len $\in$ [20-30]}} & \multicolumn{2}{c}{\textbf{len $\in$ [30-40]}} \\
\cmidrule(lr){2-3} \cmidrule(lr){4-5} \cmidrule(lr){6-7}
 \textbf{type} & \textbf{count \texttt{a}} & \textbf{count \texttt{b}} & \textbf{count \texttt{a}} & \textbf{count \texttt{b}} & \textbf{count \texttt{a}} & \textbf{count \texttt{b}} \\
\midrule
\textbf{(a)} & \colorcellgreen{86.30}{80}{100}86.30 & \colorcellgreen{86.32}{80}{100}86.20 & 62.40 & 65.20 & 50.60 & 54.40 \\
\textbf{(b)} & \colorcellgreen{90.60}{80}{100}90.60 & \colorcellgreen{94.00}{80}{100}94.00 & \colorcellgreen{80.40}{80}{100}80.40& \colorcellgreen{87.50}{80}{100}87.50& 76.10 & 79.60\\
\textbf{(c)} & \colorcellgreen{94.90}{80}{100}94.90 & \colorcellgreen{97.70}{80}{100}97.70  & \colorcellgreen{92.80}{80}{100}92.80 & \colorcellgreen{97.90}{80}{100}97.90 & \colorcellgreen{91.40}{80}{100}91.40 & \colorcellgreen{94.20}{80}{100}94.20\\
\textbf{(d)} & \colorcellgreen{93.00}{80}{100}93.00 & \colorcellgreen{94.20}{80}{100}94.20 &\colorcellgreen{87.80}{80}{100}87.80& \colorcellgreen{91.00}{80}{100}91.00& \colorcellgreen{87.30}{80}{100}87.30& \colorcellgreen{89.80}{80}{100}89.80 \\
\bottomrule
\end{tabular}

}
\caption{Counting results on strings with letter \texttt{a} and \texttt{b}, using Claude-3.5-sonnet API. All results are using supervised CoT~\cite{zhang2024supervisedchainthought}, with same prompt for GPT-4o-mini. }
\label{tab:claude}
\end{table}

Lastly, by adding the delimiter $``"$  (quotations) along with ", ", each letter will be separated into its own individual token, along with the delimiters. This simulates a case where \texttt{precise item-separated tokenization} is applied to the same counting instance.

With this string formatting design, rather than directly changing the tokenizer or applying different tokenizers to the same input, we enable the use of closed-source models, where the "tokenizer" is either unknown or cannot be directly accessed via API by users. Despite this, the desired tokenization effect can still be achieved, as the formatted strings manipulate BPE to behave as intended according to our verified assumptions.

\section{Experiments} 
\subsection{Setting}
We adopt two mainstream foundation models for our analysis of tokenization and counting ability: GPT-4o mini~\footnote{https://chatgpt.com} API and Claude-3.5-sonnet\footnote{https://claude.ai} API, considering both affordability and their representation of state-of-the-art performance. To prevent any potential cheating, we purposely disabled the use of \textit{tools} in these LLMs.
We used a Python script to generate instances for each of the four types of counting strings demonstrated in Figure \ref{fig:3}. Specifically, we generated 1000 strings under three different total length ranges: [10, 20], [20, 30], and [30, 40]. The target length was uniformly sampled within each range, and a random number of letters \texttt{a}s and \texttt{b}s were generated to form the complete string for the pure string type (Figure \ref{fig:3}.(a)). The other modified string types (Figure \ref{fig:3}.(b)-(d)) were derived by adding corresponding delimiters to the pure strings. Labels (correct counts) were generated using Python for both the number of \texttt{a}s and the number of \texttt{b}s. 

\begin{figure*}[t]
    \centering
    \includegraphics[width=\linewidth]{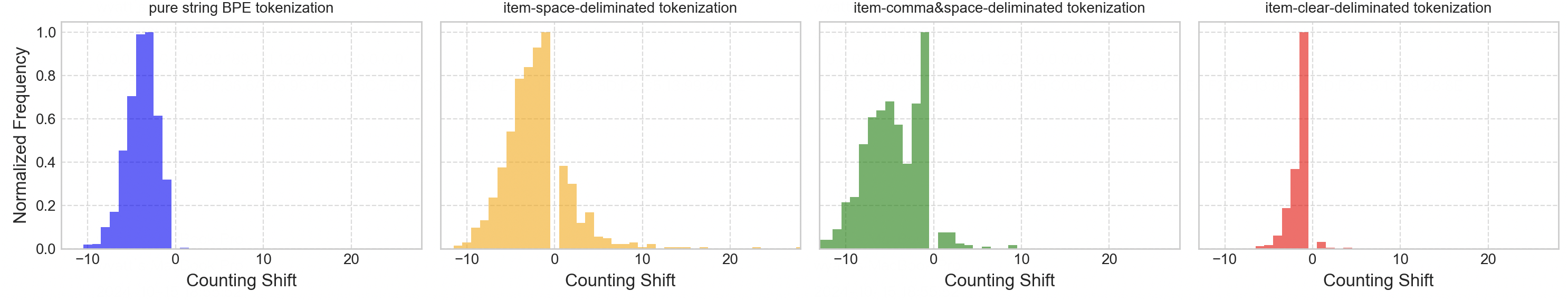}
    \vspace{-0.5em}
    \caption{Distribution of shift from correct count to GPT-calculated count, for each type of string-token fomrat (a), (b), (c) and (d) in order. The statsiticas show the results for letter \texttt{a} at length range [30, 40], as this range the error rate is high. We only calculate the shift when error is made, as correct counting instance does not have any shift.}
    \label{fig:shift}
    \vspace{-1.5em}
\end{figure*}

To eliminate potential confounding factors, such as failure to generate correct CoT steps, we use a "supervised CoT"~\cite{zhang2024supervisedchainthought} approach, where the models are explicitly instructed on what to reason at each step, i.e., the counter value. To ensure fairness, all experiments use the exact same prompt across models and conditions, as detailed in Appendix Section \ref{secCase}. The rationale behind the use of Supervised CoT, its comparison with Vanilla CoT in counting tasks, and further details are provided in Appendix Section \ref{SecSuper}.

Lastly, we evaluate the accuracy of the returned counts from the LLMs and report the average counting accuracy for each \textit{tokenization} case. The results are presented in Table \ref{tab:main_result_gpt}.
\subsection{Main Results}
We analyze the experimental results alongside our theoretical analysis of the model's upper computability, the role of CoT, and the impact of tokenization, as discussed in the previous section.
\subsubsection{CoT Grants Counting Ability}
By comparing the results for \texttt{no-CoT} and \texttt{CoT} across different length ranges and string types, we observe that CoT significantly enhances the model's counting ability in all cases, with an average performance improvement of 20\% compared to \texttt{no-CoT}.

As discussed in section 2, LMs without CoT, which rely solely on Transformer's internal latent reasoning, can only count a \textit{constant} number of times inductively. This is evident when performance drops sharply as the input length increases from [10, 20] to [30, 40] when CoT is not used. Without CoT, the counting accuracy for letters \texttt{a} and \texttt{b} falls from around 50\% to just 8\%, regardless of tokenization, which is barely above random guessing (3-4\%). However, with CoT augmentation, counting accuracy remains more stable and declines much more gradually (from 96\% drops all the way to 56\%), especially with the best tokenization choice (choice (d)). This highlights the crucial role of CoT in unlocking a model's counting potential and the practical gap from theoretical limits, caused by factors such as training quality, long-context retrieval, and CoT length constraints.

\subsubsection{Tokenization \textbf{Greatly} Affects Counting Ability}
From Table~\ref{tab:main_result_gpt}, we observe a clear and consistent trend of improved performance as tokenization choices progress from (a) to (d), indicated by increasing color darkness, when the length range is fixed. Notably, \textbf{without any} special tokenization (type (a)), counting performance is even worse than not using CoT in the same setting. For string lengths greater than 20, BPE-tokenized strings perform much closer to \textit{random guessing} than to CoT-less models. This highlights how letter-grouped BPE tokenization can \textit{severely} degrade theoretical counting ability due to the lack of token awareness.
When item-separated tokenization is applied (using delimiters such as " " or ", " in (b) and (c)), we see consistent improvements (13\%-40\%) over pure BPE, emphasizing the importance of per-item tokenization rather than grouped tokenization in counting tasks. However, since (b) and (c) still result in items being grouped with delimiters (e.g., spaces), even though the items are separated from one another, the performance limit is only reached when each item (letter) is clearly tokenized, separated from both the delimiters and other items. This eliminates the need for token-awareness, as each item-token (letter) is distinct and ready for comparison with the target, allowing the attention mechanism to function optimally (with higher cosine similarity between identical token embeddings in attention).
We conducted the same experiments with a different pair of letters, \texttt{e} and \texttt{z}, and observed identical patterns. This further demonstrates the significant impact of tokenization on counting ability, with clearly separated targets leading to markedly better results.

\subsection{Error Shifts Reveal Mistakes in Counting with BPE Tokenization}
We analyze the error shifts, defined as the difference between the model-calculated count and the true count, when mistakes occur. As shown in Figure \ref{fig:shift}, all tokenization methods result in a bias toward negative shifts, meaning the model underestimates the count. Specifically, when using pure BPE tokenization, we observe \textit{only} negative shifts. This is because BPE-tokenized tokens often contain letters the model is unaware of. For instance, the model might fail to count any "a"s in a single token like "abaa," leading to an undercount.
When item-separated tokens are introduced (as in cases (b)-(d)), positive errors begin to appear, caused by factors such as counting mistakes or long-context retrieval errors. However, with clearer token separation (case (d)), extreme shifts disappear, and the errors are mostly centered between -1 and -3, likely due to minor arithmetic mistakes by the model. This shows that while BPE incurs large extreme errors, clearer tokenization greatly mitigates counting errors.

\begin{figure}[t]
    \centering
    \includegraphics[width=0.77\linewidth]{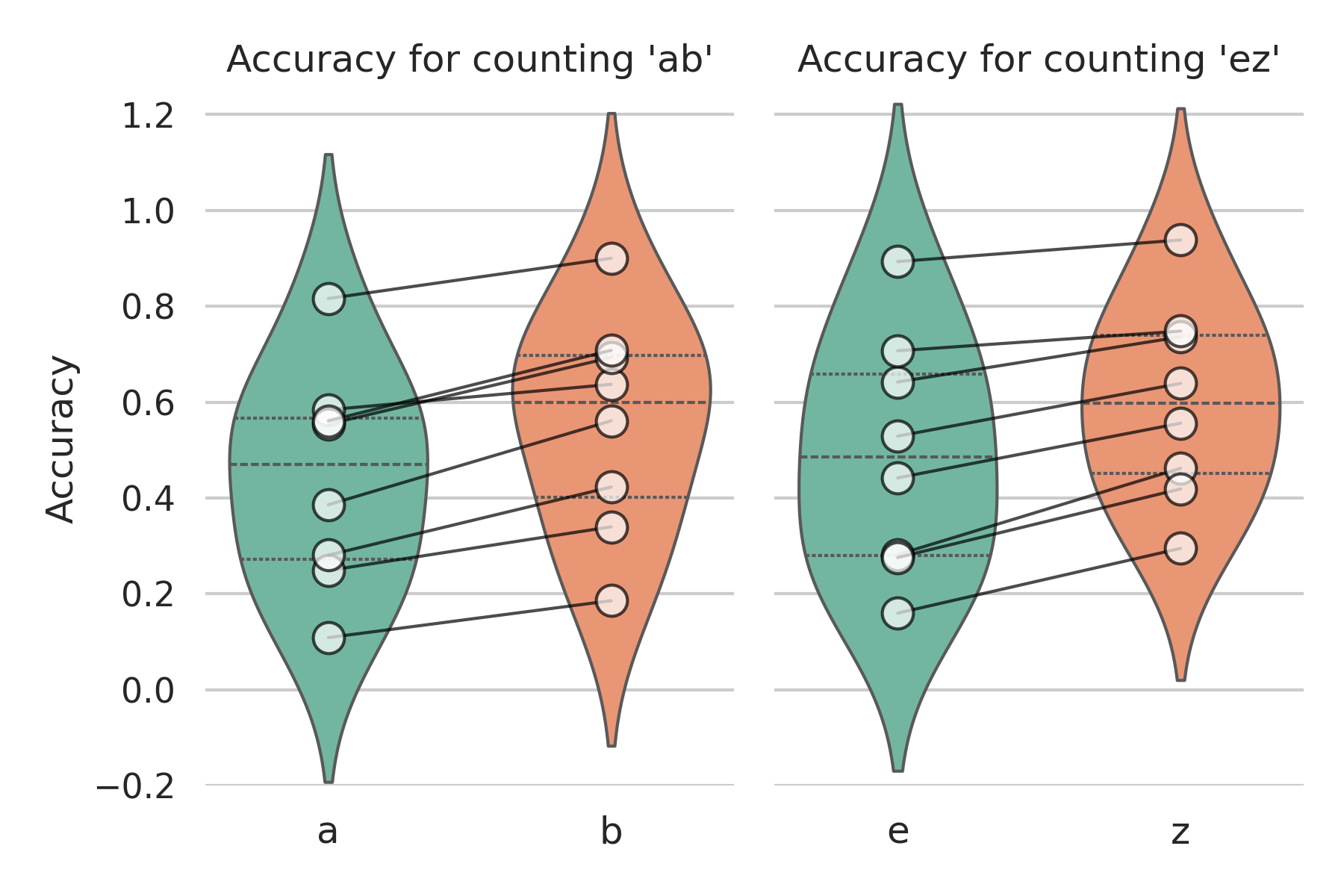}
    \vspace{-0.1em}
    \caption{Pairwise comparison of counting accuracy for different letters in strings. The left plot shows the distribution of accuracy for \texttt{a} and \texttt{b} in \texttt{ab} strings, with each dot representing the average accuracy for \texttt{a} in a given CoT case (e.g., spaced-string in the [10,20] range), connected to the corresponding accuracy for \texttt{b} in the same setting. The right plot illustrates a similar case for \texttt{e} and \texttt{z} in \texttt{ez} strings. Note: The y-axis limit exceeds [0,1] as the distribution is calculated based on variance and mean, with larger variance pushing the upper bound of the confidence interval beyond the maximum value.}
    \label{fig:violin}
    \vspace{-1em}
\end{figure}

\subsection{Different Tokens Have Varying Sensitivity in Counting Tasks}
As seen in Table~\ref{tab:main_result_gpt}, when counting is performed properly (with CoT enabled and tokenization not relying on pure BPE), we consistently observe higher accuracy when counting the letter \texttt{b} compared to the letter \texttt{a}, across all length and tokenization settings (Visualized in Figure~\ref{fig:violin} left). We suspect this discrepancy is due to differences in letter frequency in the natural language, which may affect token-embedding sensitivity in the model.
To further investigate how token frequency impacts counting accuracy in LLMs, we conducted additional counting tasks using the most frequent letter in human language,\texttt{e}, and the least frequent letter, \texttt{z}, under the same settings. As shown in Table \ref{table:ez}, we observe similar performance differences, with \texttt{z} achieving much higher counting accuracy than \texttt{e} across all proper counting settings (both with CoT and using letter-separated tokenization).
We visualize the counting performance between each pair of settings for both \texttt{ab} and \texttt{ez} in Figure~\ref{fig:violin}. The accuracy advantage of \texttt{b} over \texttt{a} and \texttt{z} over \texttt{e} is clear and significant, ranging from 3\% to 14\%. Interestingly, both \texttt{b} (1.5\%) and \texttt{z} (0.07\%) are lower-frequency tokens compared to \texttt{a} (8.2\%) and \texttt{e} (12.7\%), indicating that lower-frequency tokens are more sensitive to counting tasks in this context. We hypothesize that lower-frequency tokens contain less embedded information due to fewer occurrences during training, making them easier to identify through the attention mechanism. In contrast, frequent tokens like \texttt{a} may embed more complex information (e.g., letter \texttt{a} can be a positional noun), potentially causing distractions during attention calculations and resulting in lower performance.

\begin{figure}[ht]
    \centering
    \includegraphics[width=0.8\linewidth]{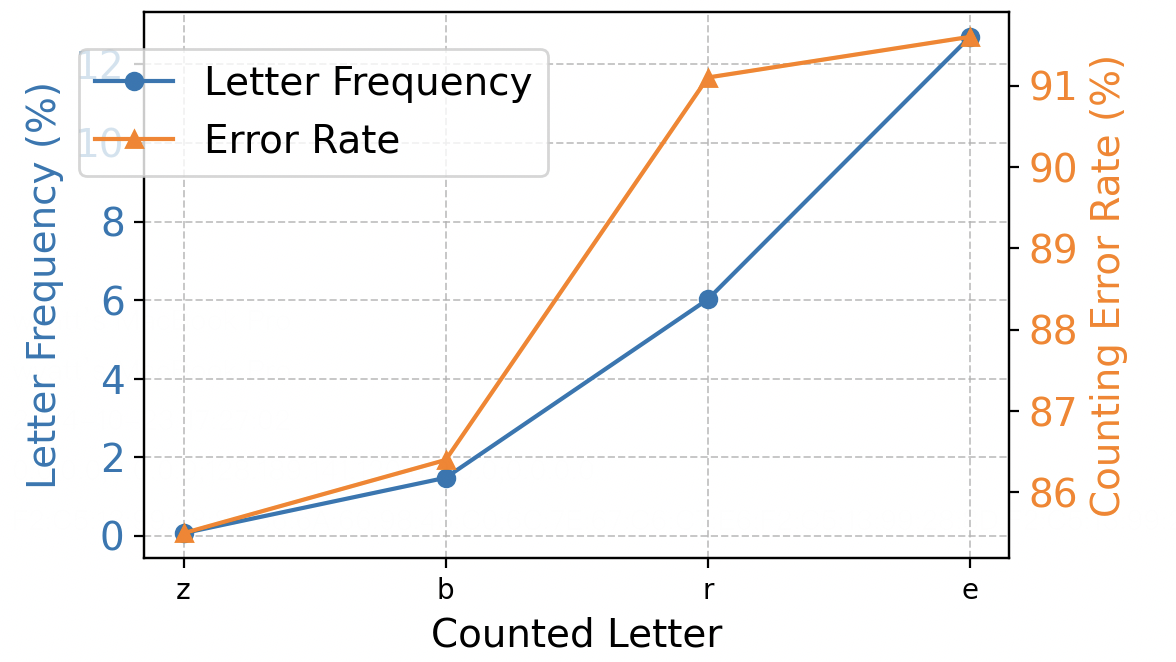}
    \vspace{-0.1em}
    \caption{Counting accuracy (Orange) with respect to target letter frequency (Blue) in Human Natural Language.}
    \label{fig:freq}
    \vspace{-1em}
\end{figure}

\noindent\textbf{Comprehensive Experiments on the Relationship Between Letter Frequency and Counting Performance.}
To verify these results beyond the letter pairs \texttt{a}, \texttt{b} and \texttt{e}, \texttt{z}, we selected another set of letters with significantly different frequencies in \textit{human languages}, according to Wikipedia: \texttt{z} (0.07\%), \texttt{b} (1.48\%), \texttt{r} (6.02\%), and \texttt{e} (12.70\%). We generated counting instances of lengths between 80 and 100—ensuring that each letter appears more than 20 times on average—by uniformly sampling one of the four letters to form each string (e.g., \texttt{zrrbeez}). We then performed counting for each letter in the generated strings. As shown in Table \ref{tab:zbre}, a consistent trend was observed across tokenization types (b)–(d) (excluding (a), as pure BPE was previously shown not to yield meaningful counting results). Specifically, rare tokens consistently outperformed more frequent tokens in natural language, with performance improvements ranging from 6\% to 12\%. Figure \ref{fig:freq} visually compares performance and letter frequency, showing an overlap between frequency and error rate. We suspect that rare letters carry less information in their embeddings, reducing distraction during the attention calculation in the counting process.

\begin{table}[t!]
\centering

\resizebox{0.4\textwidth}{!}{
\begin{tabular}{ccccc}
\toprule
\multirow{2}{*}{\textbf{String-Token Type}} & \multicolumn{4}{c}{\textbf{len $\in$ [80, 100]}} \\
\cmidrule(lr){2-5}
 & \textbf{z} & \textbf{b} & \textbf{r} & \textbf{e} \\
\midrule
\textbf{(b)} & \colorcellgreen{14.50}{8}{15}14.50 & \colorcellgreen{13.60}{8}{15}13.60 & \colorcellgreen{8.90}{8}{15}8.90 & \colorcellgreen{8.40}{8}{15}8.40 \\ \hline
\textbf{(c)} & \colorcellgreen{36}{24}{37}36.00 & \colorcellgreen{36}{24}{37}36.60 & \colorcellgreen{28}{24}{37}28.30 & \colorcellgreen{24}{24}{37}24.30 \\ \hline
\textbf{(d)} & \colorcellgreen{62}{50}{62}61.60 & \colorcellgreen{60}{50}{62}60.20 & \colorcellgreen{53}{50}{62}54.10 & \colorcellgreen{52}{50}{62}51.90 \\ \hline
\multicolumn{5}{c}{\textbf{Letter Frequency}}
\\
percentage & \colorcellgreen{1}{12}{0}0.07 & \colorcellgreen{2}{12}{0}1.48 & \colorcellgreen{6}{12}{0}6.02 & \colorcellgreen{12}{12}{0}12.70 \\ 
\bottomrule
\end{tabular}
}
\caption{Counting performance of letters that have very different letter frequency in human language. }
\vspace{-2.5em}
\label{tab:zbre}
\end{table}

\subsection{Results on Claude and Case Study with GPT-4o}
We repeated the experiments with Claude and observed similar trends, with the exception that type (c) yielded the best results, outperforming types (a)-(d), as shown in Table~\ref{tab:claude}. Upon investigation, we suspect this is because type (d) results in significantly longer CoT steps due to the higher number of tokens, leading to long-context reasoning failures in many cases for this model.
We also provide case studies using GPT-4o mini for counting tasks, including examples where CoT led to both correct and incorrect answers, as well as cases where the model predicted incorrect answers when CoT was disabled. All case studies are detailed in the Appendix Section~\ref{app:case}.

To this end, we are confident that our experimental results can be generalized to other LLMs, given that the training methods and tokenization strategies (as demonstrated in Appendix Section \ref{app:token}) are nearly identical, leading to counting being performed in a similar manner across such models.



\section{Conclusion}
This study has demonstrated that tokenization plays a significant role in the reasoning capabilities of large language models (LLMs), particularly in tasks such as counting. The way input data is tokenized influences how models process and interpret information, ultimately affecting performance outcomes. Therefore, careful consideration must be given to the selection of tokenization strategies when working with LLMs, especially in tasks where fine-grained reasoning is essential. Our findings suggest that the exploration and optimization of tokenization methods is not only relevant but crucial for improving model performance. Future research should further investigate the impact of different tokenization techniques, with a focus on refining these methods to enhance LLM reasoning abilities across a broader range of tasks.

\section*{Limitations}
Our experiments were conducted on GPT-4o and Claude-3.5, and while both models demonstrated strong patterns and consistent evidence showing that certain types of tokenization significantly improve counting performance, we did not extend our testing to other LLMs such as LLaMA or Mistral. This was primarily due to budget and time constraints, as well as preliminary findings that these models exhibited weaker instruction-following abilities compared to GPT and Claude, making the evaluation process more challenging. However, we believe our research remains robust despite these limitations, as mainstream model training and design principles are largely universal, and the patterns observed are likely generalizable to other LLMs.

Additionally, our experiments did not explore extreme context lengths, such as counting instances with more than several hundred tokens. We found that such cases often led to instability due to the accumulation of long CoT steps. We aim to further investigate this aspect as LLMs improve in handling long-context retrieval and generation.
\bibliography{custom}
\bibliographystyle{acl_natbib}

\appendix

\section*{Appendix}
\section{Supervised Chain of Thought}
\label{SecSuper}
Naive Chain of Thought (CoT), which uses a generic "think step by step" prompt for all tasks, poses significant challenges for models in determining the correct steps, especially for complex, multi-step reasoning tasks. To mitigate this confounding factor, we follow previous work and employ Supervised CoT~\cite{zhang2024supervisedchainthought}, as the derivation of steps is not the focus of our research and should not affect performance due to incorrect CoT steps. Below, we define Supervised CoT and explain its application in counting tasks.
\subsection{Definition}
The search space for solving a task can be viewed as a combination of the prompt space and the answer space. When instructed to perform tasks step by step, language models must devise a step template which is used to determine the actions at each step. This template is crucial for solving tasks, as it specifies what information is processed and how it is computed at each CoT step. However, for a given task, there are numerous ways to perform a "step-by-step" approach, each computing different elements per step. Finding the optimal set of steps is challenging yet essential, as it directly influences the ability to find solutions in the answer space~\cite{zhang2024supervisedchainthought}.

Supervised CoT provides human supervision in determining the step template. Rather than asking the model to develop its own plan for each step, humans identify the "recurrent" procedure in the computation and explicitly instruct the model to follow a specific step template. This approach allows the CoT to bypass the need to search for optimal steps, focusing instead on finding solutions within the answer space under optimal step guidance.
\subsection{Supervised CoT and Counting}
In inductive counting, which relies on CoT to compute the counter value recurrently(Figure \ref{fig:inductive}), it is crucial that each step of CoT accurately extracts and outputs the counter value in text. This output is necessary for the value to be recurrently processed through "string-vector" conversion. Therefore, rather than simply prompting the model with "determine the number of \texttt{a} in the given string" using the generic instruction "think step by step," we specifically instruct the model to print out a counter value at each step. We explicitly define the step template to ensure the model follows the optimal CoT steps, preventing deviations or the use of suboptimal steps.

\begin{table*}[ht]
\centering

\resizebox{0.7\textwidth}{!}{
\begin{tabular}{ccc|cc}
\toprule
\multirow{2}{*}{\textbf{String-token Type}} & \multicolumn{2}{c|}{\textbf{Counting \texttt{a}}} & \multicolumn{2}{c}{\textbf{Counting \texttt{b}}} \\
\cmidrule(lr){2-3} \cmidrule(lr){4-5}
 & \textbf{Unsupervised-CoT} & \textbf{Supervised CoT} & \textbf{Unsupervised-CoT} & \textbf{Supervised CoT} \\
\midrule
\textbf{(b)} & 8.40 & \textbf{10.90} & \textbf{20.70} & 18.60 \\
\textbf{(c)} & 24.00 & \textbf{28.10} & 29.30 & \textbf{42.30} \\
\textbf{(d)} & 34.90 & \textbf{56.10} & 42.70 & \textbf{70.80} \\

\bottomrule
\end{tabular}
}
\caption{Counting experiments in the length range of 30-40 comparing Supervised CoT and Unsupervised CoT. The bolded font indicates the better performance in the pairwise comparison between Supervised and Unsupervised CoT.}
\label{tab:Supervision}
\end{table*}
\noindent\textbf{Experiments.} We demonstrate the significant performance gap between Supervised and Unsupervised CoT. Specifically, we observe that supervision not only helps the model accurately extract the counter but also ensures it follows the correct steps (e.g., an incorrect step would be outputting whether the current letter is the target, rather than extracting the counter value). Even when Unsupervised CoT identifies the correct steps (i.e., extracting the counter into text), we still notice more frequent errors during the extraction process compared to Supervised CoT, which imposes strict constraints on what to extract at each step. The comparison between Supervised and Unsupervised CoT is presented in Table \ref{tab:Supervision}, showing a clear dominance of Supervised CoT, with accuracy gains observed in nearly all cases.

\section{Replication Experiments Note}
We have open-sourced the experimental results for every instance of each experiment, in the provided GitHub link, to facilitate future research and analysis by other researchers. All reported experiment numbers are stable, using the same experimental settings and prompts. Specifically, we observe an average variance in accuracy of less than 1\% across runs of the same experiments, indicating that they are fully replicable with the same GPT-4o version used. However, note that updates to the API version may cause potential variations in results, which are beyond our control.

\section{Tokenization in Different LLMs}
\begin{figure*}[t]
    \centering
    \includegraphics[width=\linewidth]{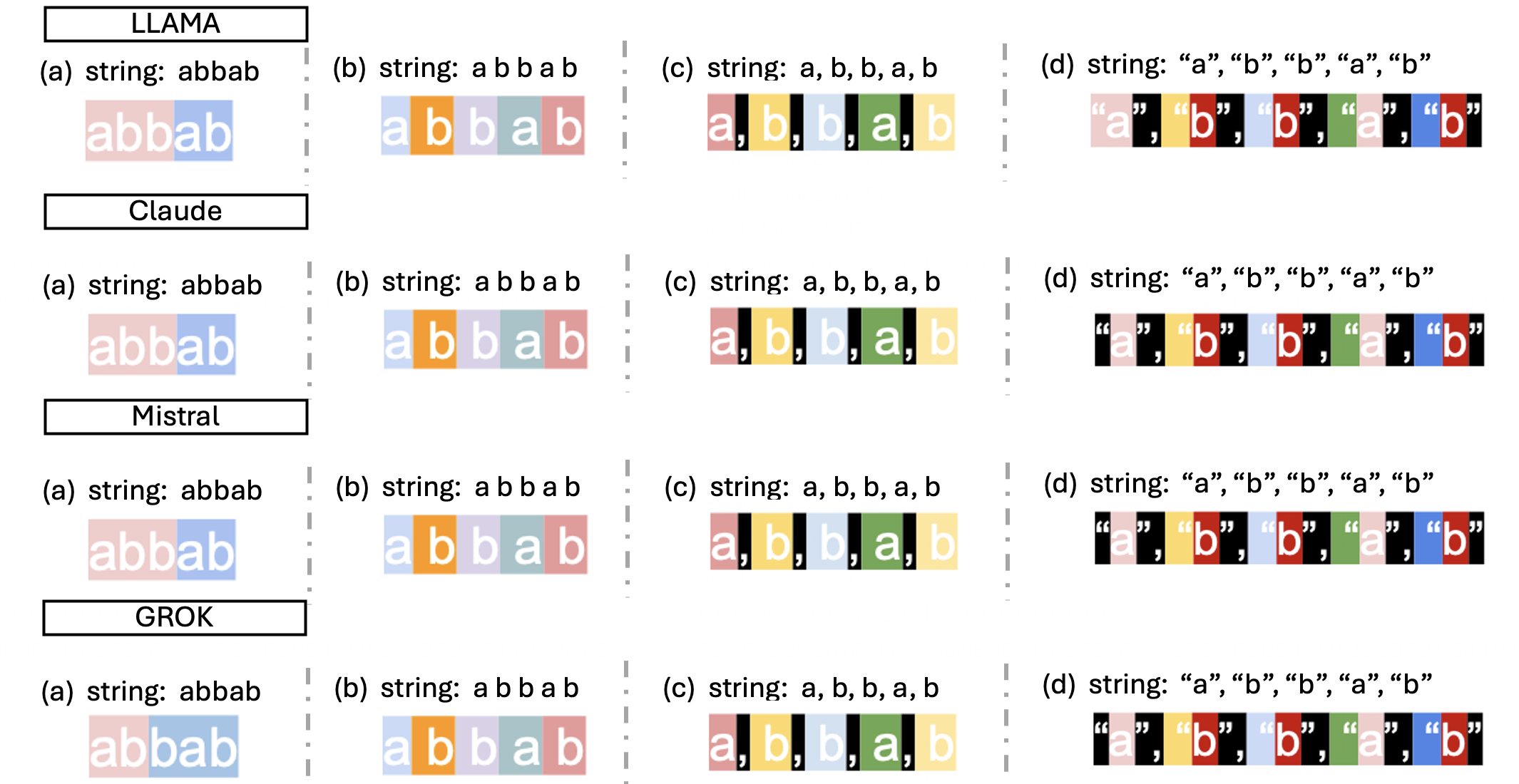}
    \vspace{-0.5em}
    \caption{Difference in tokenization on strings when counting instances are presented in different formats, across different LLMs. }
    \label{fig:tokenall}
    \vspace{-1.5em}
\end{figure*}

Figure \ref{fig:tokenall} illustrates the tokenization of input strings across various LLMs. We investigate both language models and multi-modal models, observing nearly identical tokenization behaviors across all tested models. Specifically, pure strings (e.g., \texttt{abbab}) may be tokenized differently due to the varying byte pair encoding (BPE) schemes used by each tokenizer. When a token-level delimiter is introduced, we consistently observe that the delimiter is combined with the adjacent letter, aligning with our previous assumptions. Additionally, some models handle the initial token differently, resulting in the first letter being treated as a standalone token (e.g., in the Grok model) or being combined with a quotation delimiter (e.g., in GPT-4o).
In summary, our string design effectively allows us to manipulate modern LLMs to tokenize identical counting instances into different, desired tokens.
\label{app:token}
\section{Case Study}
\label{secCase}
Tables \ref{tab:casestudy1}-\ref{tab:casestudy4} present cases for each type of token counted using CoT. As shown, Supervised CoT consistently adheres to a strict step-by-step template, accurately extracting the counter at each step. In contrast, Unsupervised CoT often skips crucial steps or deviates from the optimal method of extracting counters. Additionally, Supervised CoT with type (d) tokenization tends to produce much longer reasoning contexts, yet still achieves the best performance due to the combination of optimal tokenization and supervision.
\label{app:case}

\label{sec:appendix}

\begin{table*}[htbp]
    \centering
    \small\vspace{-.2cm}
    \resizebox{0.9\textwidth}{!}
    {%
    \begin{tabular}{|p{1\linewidth}|}

    \hline
      \textbf{Prompt:}\\ Task: Count the number of occurrences of the substring 'a' in the given string step by step. Do not use bold font in the response.

Instructions:

    Iterate through the string and count each occurrence of the substring.
    Use a counter to keep track of how many times the substring appears.
    Output your step-by-step counting process.
    Conclude with: 'Result: ' followed by the number.

String: ["b", "b", "a", "a", "b". "b", "a", "b", "b", "b", "a", "a", "b", "a", "b", "a", "a", "b", "b", "a", "b"]

Response: \\ \hline
      \textbf{Supervised CoT Result (Experiment 4):} 
    \\ "1. Initialize a counter to 0.2. Start iterating through the string.3. Check the first element 'b'. It does not match 'a', so the counter remains 0.4. Move to the second element 'b'. It does not match 'a', so the counter remains 0.5. Check the third element 'a'. It matches 'a', so increment the counter to 1.6. Move to the fourth element 'a'. It matches 'a', so increment the counter to 2.7. Check the fifth element 'b'. It does not match 'a', so the counter remains 2.8. Move to the sixth element 'b'. It does not match 'a', so the counter remains 2.9. Check the seventh element 'a'. It matches 'a', so increment the counter to 3.10. Move to the eighth element 'b'. It does not match 'a', so the counter remains 3.11. Check the ninth element 'b'. It does not match 'a', so the counter remains 3.12. Move to the tenth element 'b'. It does not match 'a', so the counter remains 3.13. Check the eleventh element 'a'. It matches 'a', so increment the counter to 4.14. Move to the twelfth element 'a'. It matches 'a', so increment the counter to 5.15. Check the thirteenth element 'b'. It does not match 'a', so the counter remains 5.16. Move to the fourteenth element 'a'. It matches 'a', so increment the counter to 6.17. Check the fifteenth element 'b'. It does not match 'a', so the counter remains 6.18. Move to the sixteenth element 'a'. It matches 'a', so increment the counter to 7.19. Check the seventeenth element 'a'. It matches 'a', so increment the counter to 8.20. Move to the eighteenth element 'b'. It does not match 'a', so the counter remains 8.21. Check the nineteenth element 'b'. It does not match 'a', so the counter remains 8.22. Move to the twentieth element 'b'. It does not match 'a', so the counter remains 8.23. Check the twenty-first element 'a'. It matches 'a', so increment the counter to 9.24. Move to the twenty-second element 'b'. It does not match 'a', so the counter remains 9. Result: 9\\ \hline
    \textbf{Correct Answer:}\\9 \\ \hline
    \end{tabular} }\vspace{-.1cm}
    \caption{Case Study 1, Correct Prediction with Supervised Chain-of-Thought}\vspace{-.2cm}
    \label{tab:casestudy1}
\end{table*}

\begin{table*}[htbp]
    \centering
    \small\vspace{-.2cm}
    \resizebox{0.9\textwidth}{!}
    {%
    \begin{tabular}{|p{1\linewidth}|}

    \hline
      \textbf{Prompt:}\\ Count the number appearances of 'a's in the string below. Directly output 'Result: ' followed by the counted number. Do not use bold font in the response.

String: ["b", "b", "a", "a", "b". "b", "a", "b", "b", "b", "a", "a", "b", "a", "b", "a", "a", "b", "b", "a", "b"]

Response:\\ \hline
      \textbf{Result (Experiment 4):} \\ Result: 7\\ \hline\textbf{Correct Answer:}\\9 \\ \hline
    \end{tabular} }\vspace{-.1cm}
    \caption{Case Study 2, Incorrect Prediction Prompting Banning the usage of Chain-of-Thought. }\vspace{-.2cm}
    \label{tab:casestudy2}
\end{table*}

\begin{table*}[htbp]
    \centering
    \small\vspace{-.2cm}
    \resizebox{0.9\textwidth}{!}
    {%
    \begin{tabular}{|p{1\linewidth}|}

    \hline
      \textbf{Prompt:}\\  Determine the number of appearances of '\{substring\}'s in the string below. Think step by step. Directly output 'Result: ' followed by the counted number. Do not use bold font in the response.

String: b b a a b b b a b b b b b a a a b b b b

Response:\\ \hline
      \textbf{Result (Experiment 2):} \\ To determine the number of appearances of 'a's in the given string, we will analyze the string step by step. 1. Let's first isolate the string and count the occurrences of the letter 'a'.      The given string is:   "b b a a b b b a b b b b b a a a b b b b" 2. We can count the 'a's by looking through the string or splitting it into individual elements to count:   - The string has the following elements: b, b, a, a, b, b, b, a, b, b, b, b, b, a, a, a, b, b, b, b.   3. Now we count each 'a':   - First occurrence: a (1)   - Second occurrence: a (2)   - Third occurrence: a (3)   - Fourth occurrence: a (4)  - Fifth occurrence: a (5) 4. Total occurrences of 'a': 5 Result: 5\\ \hline \textbf{Correct Answer:}\\6\\ \hline
    \end{tabular} }\vspace{-.1cm}
    \caption{Case Study 3, Incorrect Prediction with Unsupervised Chain-of-Thought}\vspace{-.2cm}
    \label{tab:casestudy3}
\end{table*}

\begin{table*}[htbp]
    \centering
    \small\vspace{-.2cm}
    \resizebox{0.9\textwidth}{!}
    {%
    \begin{tabular}{|p{1\linewidth}|}

    \hline
      \textbf{Prompt:}\\  Task: Count the number of occurrences of the substring 'a' in the given string step by step. Do not use bold font in the response.

Instructions:

    Iterate through the string and count each occurrence of the substring.
    Use a counter to keep track of how many times the substring appears.
    Output your step-by-step counting process.
    Conclude with: 'Result: ' followed by the number.

String: bbaabbbabbbbbaaabbbb

Response: \\ \hline
      \textbf{Result (Experiment 1):} \\ Let's go through the string "bbaabbbabbbbbaaabbbb" step by step to count the occurrences of the substring 'a'. 1. Initialize a counter to 0. 2. Start reading the string from the first character. - Character 1: 'b' (no 'a', counter remains 0) - Character 2: 'b' (no 'a', counter remains 0) - Character 3: 'a' (found 'a', increment counter to 1) - Character 4: 'a' (found 'a', increment counter to 2) - Character 5: 'b' (no 'a', counter remains 2) - Character 6: 'b' (no 'a', counter remains 2) - Character 7: 'b' (no 'a', counter remains 2) - Character 8: 'a' (found 'a', increment counter to 3) - Character 9: 'b' (no 'a', counter remains 3) - Character 10: 'b' (no 'a', counter remains 3) - Character 11: 'b' (no 'a', counter remains 3) - Character 12: 'b' (no 'a', counter remains 3) - Character 13: 'a' (found 'a', increment counter to 4) - Character 14: 'a' (found 'a', increment counter to 5) - Character 15: 'b' (no 'a', counter remains 5) - Character 16: 'b' (no 'a', counter remains 5) - Character 17: 'b' (no 'a', counter remains 5) - Character 18: 'b' (no 'a', counter remains 5) 3. End of the string reached. Result: 5 \\ \hline \textbf{Correct Answer:}\\6 \\ \hline
    \end{tabular} }\vspace{-.1cm}
    \caption{Case Study 4, Incorrect Prediction with Supervised Chain-of-Thought}\vspace{-.2cm}
    \label{tab:casestudy4}
\end{table*}

\end{document}